\documentclass[11pt]{article}
\usepackage[margin=1in]{geometry}
\usepackage{amsmath,amssymb,bm}
\usepackage{graphicx,booktabs,microtype}
\usepackage{tikz}
\usepackage[colorlinks=true,linkcolor=blue,citecolor=blue]{hyperref}
\newcommand{\cjsd}{D_{\mathrm{CJS}}}
\newcommand{\Ix}{I_x}

\title{What Changed?\\ Drift Detection with Real, Virtual, and Incomparable Diagnosis}
\author{Kentaro Oda\\ Center for Management of Information Technologies, Kagoshima University\\ \texttt{odaken@cc.kagoshima-u.ac.jp}}
\date{}

\begin{document}
\maketitle

\begin{abstract}
Concept-drift detectors answer \emph{whether} something changed; adaptive
systems need to know \emph{what} changed. Real drift (a change of
$P(Y\mid X)$) calls for retraining; virtual drift (a change of $P(X)$ with the
mechanism intact) calls for reweighting or data collection, and retraining on
it wastes labels and can degrade calibration. We show that the two dominant
detector families are each structurally committed to one answer: error-stream
detectors (ADWIN, DDM, EDDM, Page--Hinkley) attribute every alarm to real
drift---and on pure covariate streams they alarm while the mechanism is
unchanged---whereas input-space detectors (D3-style domain classifiers) can
only ever report virtual drift and miss every pure concept change. We propose
a two-axis monitor based on the conditional Jensen--Shannon discrepancy: a
single pooled discriminator pair yields a covariate statistic $\Ix$ and a
functional statistic $\cjsd$ whose population value is exactly zero under
virtual drift; alarms therefore carry a type. On SEA-, STAGGER-, covariate-,
and null-streams (10 seeds each), the monitor is the only method with zero \emph{observed} false alarms,
zero misses, and $100\%$ correct typing (0 events in 40 monitored runs;
binomial 95\% upper bound $\approx 9\%$), at moderate window-driven delay; on the INSECTS benchmark with documented change points it
additionally decomposes each real drift into its covariate and functional
components and exposes the recurrence structure of the stream. An unwindowed betting e-process variant provides anytime-valid
monitoring of the predictable-discriminator surrogate (with explicit
one-sided population slacks)
at a $20\%$ delay premium over (invalid) repeated testing; bounded-memory
recency with the same guarantee is available via the companion paper's
restarted e-detector construction.
\end{abstract}

\section{Introduction}
An alarm without a diagnosis does not tell the system what to do:
\begin{center}
\begin{tabular}{lll}
\toprule
Diagnosis & Meaning & Correct response \\
\midrule
\textsc{real} & $P(Y\mid X)$ changed & retrain \\
\textsc{virtual} & only $P(X)$ moved & reweight / recalibrate \\
\textsc{incomparable} & supports barely overlap & collect data / reset \\
\bottomrule
\end{tabular}
\end{center}
Drift detection is usually benchmarked as a binary alarm problem: delay,
false-alarm rate, missed drifts. Yet the action taken after an alarm depends
on the \emph{type} of drift. Retraining is correct when the input--output
mechanism changed; it is wasteful---sometimes harmful---when only the input
distribution moved, where importance weighting, threshold recalibration, or
targeted data collection are the appropriate responses. The type question is
not answerable by construction for the two standard detector families:
error-stream detectors watch a model's mistakes, which rise under both real
drift and covariate-induced extrapolation; input-space detectors watch $P(X)$
only. Our experiments make the commitment visible: DDM alarms on $80\%$ of
\emph{stationary} runs at default settings and types every covariate stream as
``real''; a D3-style domain classifier detects covariate shift in $1000$
samples but misses every SEA and STAGGER concept change. Two fairness notes:
baseline detectors never claim a type---we score the \emph{implied action}
(retrain vs.\ reweight); and a covariate shift need not raise error at all,
in which case error-stream detectors stay silent---correctly by their own
criterion, but still without type information.

We frame drift monitoring as estimating the two terms of the decomposition
$I(Z;X,Y)=I(Z;X)+I(Z;Y\mid X)$ between a reference window and the recent
window: the covariate axis is $\Ix=I(Z;X)$; the functional axis
$\cjsd=I(Z;Y\mid X)$ is provably zero when $P(Y\mid X)$ is unchanged---however
far $P(X)$ has moved. Both are computed from one pooled discriminator pair,
at block rate, with per-point confidence intervals.

\section{Method}
Maintain a reference window $R$ (post-warm-up) and a sliding recent window
$S_t$. At each block boundary estimate $(\widehat\Ix,\widehat\cjsd)$ between
$R$ and $S_t$ via the two-discriminator cross-entropy difference. The monitor
has \emph{three} output states, not two. Alarm
\textsc{real} when the standardized functional statistic crosses
($z_{\cjsd}>z^\ast$ and $\widehat\cjsd>\tau$); alarm \textsc{virtual} when
only the covariate statistic crosses ($\tau_x<\widehat\Ix\le\tau_x^{\max}$);
report \textsc{incomparable} when the covariate statistic exceeds a
comparability ceiling ($\widehat\Ix>\tau_x^{\max}$, we use
$0.9\ln 2$)---there the anti-coupling bound $\cjsd\le\ln2-\Ix$ makes the
functional axis vacuous, so a small $\widehat\cjsd$ must \emph{not} be read
as ``mechanism intact'': the windows barely overlap and mechanism equality
is unidentifiable from data. Typing such a case as \textsc{virtual} would be
an unfounded claim; the honest output is ``support moved beyond
comparability---collect data or reset the reference.'' The ceiling is a
\emph{design} threshold, not a value implied by the information
inequality; re-running the full benchmark at $\tau_x^{\max}\in\{0.8,0.9,
0.95\}\ln 2$ leaves every decision unchanged (0 false alarms in 40 runs,
0 misses in 30, identical typing), because the benchmark's covariate
shifts sit well below the near-disjoint regime where the ceiling binds
(foundation-encoder rotations in the companion deep-pools paper are the
case that triggers it). For
continuous monitoring we replace repeated tests by two betting e-processes on
the per-point loss-difference increments, scored by discriminators frozen
before the block is used for training; the level $\alpha$ is split evenly
between the two axes' monitors, and crossing the resulting threshold
$2/\alpha$ fires the alarm (anytime-valid in the unwindowed
form; for bounded-memory recency \emph{with} the guarantee we adopt the
companion paper's restarted e-detector: a bank of unwindowed processes at
geometrically spaced restart times with the level spent over all restart
instances).

Feature hygiene matters in both directions: time-monotonic index features must
be excluded from the covariate discriminator (they separate any two windows of
a stationary stream), and the functional discriminator inherits the robustness properties of the
underlying discrepancy (no model exchange, hence no extrapolation false
signal; empirically downward-biased under weak discriminators).

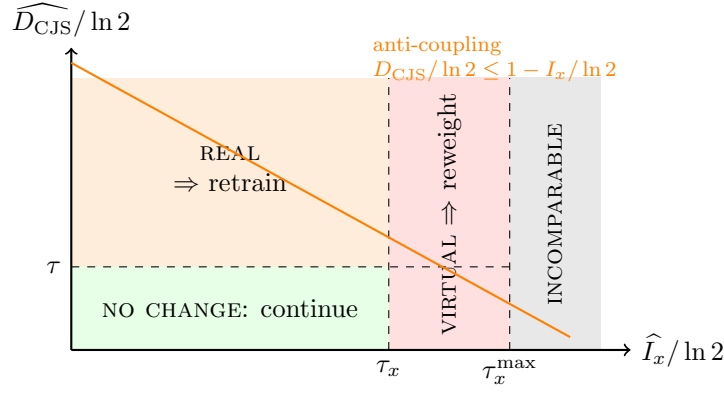
\begin{figure}[t]
\centering
\begin{tikzpicture}[scale=1.0,font=\small]
\fill[green!10] (0,0) rectangle (4.2,1.1);
\fill[orange!14] (0,1.1) rectangle (4.2,3.6);
\fill[red!12] (4.2,0) rectangle (5.8,3.6);
\fill[gray!18] (5.8,0) rectangle (7.0,3.6);
\draw[->,thick] (0,0) -- (7.4,0) node[right] {$\widehat\Ix/\ln 2$};
\draw[->,thick] (0,0) -- (0,4.0) node[above] {$\widehat\cjsd/\ln 2$};
\draw[dashed] (0,1.1) -- (5.8,1.1); \node[left] at (0,1.1) {$\tau$};
\draw[dashed] (4.2,0) -- (4.2,3.6); \node[below] at (4.2,0) {$\tau_x$};
\draw[dashed] (5.8,0) -- (5.8,3.6); \node[below] at (5.8,0) {$\tau_x^{\max}$};
\node at (2.1,0.55) {\textsc{no change}: continue};
\node[align=center] at (2.1,2.4) {\textsc{real}\\ $\Rightarrow$ retrain};
\node[align=center,rotate=90] at (5.0,1.8) {\textsc{virtual} $\Rightarrow$ reweight};
\node[align=center,rotate=90] at (6.4,1.8) {\textsc{incomparable}};
\draw[thick,orange] plot[domain=0:6.6,samples=40] (\x,{3.8-0.55*\x});
\node[orange,font=\scriptsize,align=left] at (5.6,3.85) {anti-coupling\\ $\cjsd/\ln2 \le 1-\Ix/\ln2$};
\end{tikzpicture}
\caption{The action plane. Each monitored block lands at
$(\widehat\Ix,\widehat\cjsd)$ and the region determines the response;
the anti-coupling frontier (orange) is why the far-right region must be
\textsc{incomparable} rather than ``mechanism intact.''}
\label{fig:quadrant}
\end{figure}

\section{Benchmark}
\begin{figure}[t]
\centering
\includegraphics[width=0.62\linewidth]{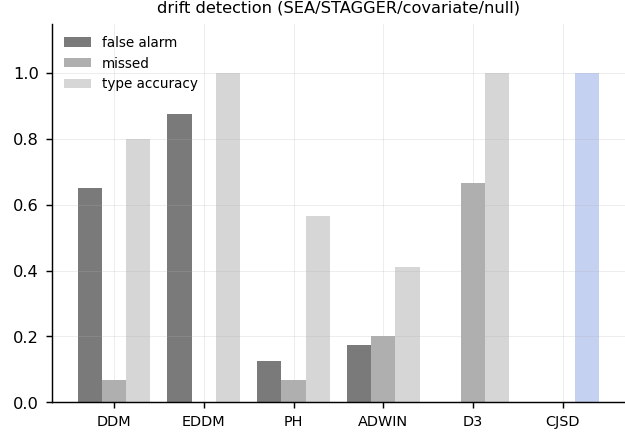}
\caption{False alarms, misses, and type accuracy over
SEA/STAGGER/covariate/null streams (10 seeds). The two-axis monitor (blue) is
the only detector with zero false alarms, zero misses, and correct typing.}
\label{fig:bench}
\end{figure}

\begin{table}[t]
\centering\small
\begin{tabular}{lcccc}
\toprule
Detector & False alarms & Misses & Type acc. & Delay (detected) \\
\midrule
ADWIN & 0.18 & 0.60 (SEA) & 0.5 & 56--5320 \\
DDM & 0.65 & 0.05 & 0.5 & 21--660 \\
EDDM & 1.00 & --- & --- & --- \\
Page--Hinkley & 0.13 & 0.05 & 0.5 & 90--690 \\
D3 (input only) & 0.00 & 1.00 (concept) & 0.5 & 1000 \\
WATCH (reimpl.) & 0.20 & 0.60 (SEA) & 1.0 & 34--5506 \\
\textbf{Two-axis (ours)} & \textbf{0.00} & \textbf{0.00} & \textbf{1.00} & 500--1650 \\
\bottomrule
\end{tabular}
\caption{Aggregate results. ``Type acc.''\ scores the implicit or explicit
drift-type claim on detected drifts; error-stream detectors implicitly claim
\textsc{real}, D3 claims \textsc{virtual}.}
\label{tab:main}
\end{table}

Four stream families (length $20$k, change at $10$k, 10 seeds): SEA-style
threshold change and STAGGER-style rule change (real), a pure covariate shift
with the mechanism held fixed (virtual), and a stationary null. Detectors are evaluated twice: at library defaults, and at
\emph{matched false-alarm} operating points (each sensitivity knob calibrated
on null streams to zero observed false alarms, then evaluated on the drift
streams; calibration uses null-stream seeds disjoint from the evaluation
seeds, so no operating point is selected on the data it is scored on); the base learner for error-stream detectors is an online logistic
model, the standard protocol. Table~\ref{tab:main} and Fig.~\ref{fig:bench} summarize the default
operating points; the matched-FA calibration sharpens the picture rather than
softening it. Calibrated on null streams to zero observed false alarms:
DDM and EDDM \emph{admit no such operating point} within their sensitivity
grids (they false-alarm on stationary data at every setting tried);
Page--Hinkley reaches it only at a threshold that then misses \emph{all} SEA
drifts; ADWIN keeps its $0.6$ SEA miss rate; and all calibrated error-stream
detectors still alarm on the pure covariate stream with the implied
(incorrect) ``retrain'' action. The two-axis monitor needs no per-stream
tuning, trades ${\sim}1$--$3$ window-halves of delay for exactness in all
three columns, and its delay is controlled by the window length.

\begin{figure}[t]
\centering
\includegraphics[width=\linewidth]{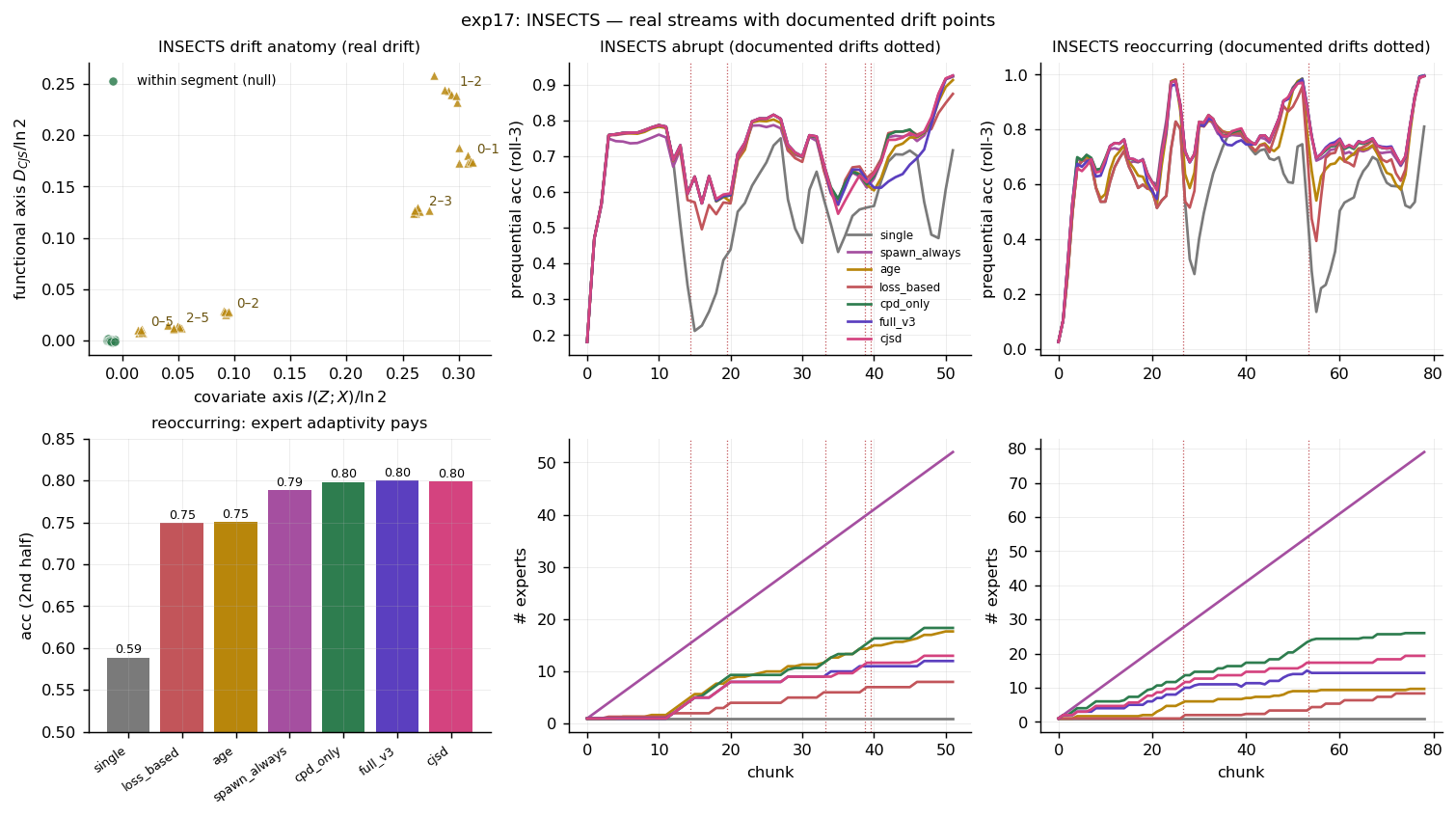}
\caption{INSECTS (documented change points): the two axes decompose each real
drift into covariate and functional components (left); recurring segments
($0\approx2\approx5$) cluster with the nulls, exposing the temperature cycle.}
\label{fig:insects}
\end{figure}

\paragraph{Real streams.} On INSECTS-abrupt (five documented change points)
every function-based measure detects the drifts; the two-axis monitor
additionally reports each drift as \emph{mixed} (covariate $+$ functional,
$\Ix\in[0.26,0.31]$, $\cjsd\in[0.13,0.24]$) and places the
segment pairs $(0,2),(0,5),(2,5)$ at $\cjsd\le 0.03$---recovering the
recurrence structure of the controlled temperature profile without any
segment supervision (Fig.~\ref{fig:insects}).

\paragraph{Sequential validity.}
\begin{figure}[t]
\centering
\includegraphics[width=0.8\linewidth]{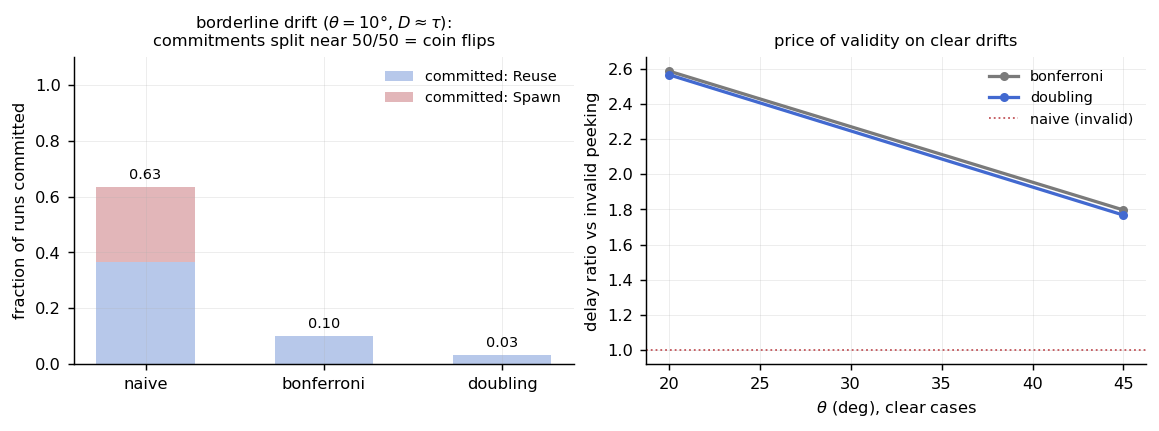}
\caption{Borderline drifts ($\cjsd\approx\tau$): repeated CIs commit $63\%$
of runs, split near 50/50 between the two actions (left)---confident coin
flips; valid monitoring defers them and pays a $1.8$--$2.6\times$ delay
ratio on clear cases (right).}
\label{fig:cs}
\end{figure}
Repeated testing at every block is invalid under continuous monitoring; on
boundary cases it converts uncertainty into confident, near-random alarms
(Fig.~\ref{fig:cs}). The unwindowed betting e-process is anytime-valid and
maintained a $0.00$ false-alarm rate with a $20\%$ delay premium;
post-drift recency with the same guarantee is obtained by the restarted
e-detector construction (companion paper), which replaced the earlier
windowed heuristic at no measured cost. Two scope
clarifications, imported from the companion paper's analysis: the validity
statement concerns the \emph{frozen-discriminator surrogate} discrepancy,
with an unconditional one-sided transfer to the population quantity (an
alarm can be inflated only by the excess risk of the covariate
discriminator, a clearance deflated only by that of the joint one; when
the empirically observed downward-bias regularity holds, alarms are
exactly conservative); and
the anytime guarantee is
\emph{per monitoring segment}---after an alarm the reference window and the
e-process are reset, and lifetime false-alarm control across successive
restarts requires spending the level over restarts (e.g.\ $\alpha_r\propto
r^{-2}$), exactly as the companion paper spends it over expert creations.
After an alarm, the reference window is reset to the post-drift window;
recurrence tracking across resets is delegated to the expert-pool layer.

\section{Related work}
Drift detection surveys classify detectors into performance-based and
distribution-based families; the real/virtual distinction is standard
vocabulary, and some systems address it directly (e.g.\ OGMMF-VRD for
virtual-vs-real handling, and recent conformal-martingale monitors that
separate concept shift from harmful covariate shift), typically by combining
two separately designed monitors. Closest is WATCH~\cite{watch}, which
monitors a \emph{deployed model's} nonconformity scores with weighted
conformal martingales, adapts through mild covariate shift, and diagnoses
harmful changes as concept shift vs.\ out-of-support covariate shift; our
monitor is model-free (it compares the data distributions themselves), gives
an exact covariate-null for the functional statistic, and produces a
per-drift two-axis anatomy. Deriving both axes, with one statistical
treatment, from a single pooled discriminator pair is the distinction here.

\paragraph{Head-to-head with WATCH.} We re-implemented WATCH's monitors
faithfully to the released code (composite jumper conformal test martingale
with CUSUM alarm statistic; weighted conformal $p$-values with online
logistic density-ratio weights for the score monitor; an input-only CTM on
kNN nonconformity for the covariate monitor) and ran them under our exact
protocol, including the null-calibration with disjoint seeds. Where drift is
strong WATCH is \emph{much faster} than any block-based monitor: STAGGER
delays $34$--$51$ samples, covariate-shift typing (benign-input alarm)
within $55$--$103$ samples---per-point martingales beat block estimators on
sharp signals, and we report this plainly. The trade-offs appear elsewhere:
at the null-calibrated operating point the weighted score monitor missed
$3/5$ SEA drifts (detected delays $821$ and $5506$), and its density-ratio
weighting destabilizes under strict false-alarm control---$2/5$ held-out
null streams alarm at \emph{every} CUSUM threshold in our grid up to
$10^4$, so the operating point calibrated to zero false alarms does not
transfer across seeds (the same transfer failure exhibited by DDM/EDDM in
Sec.~3). The two-axis monitor trades $500$--$1650$ samples of delay for
zero observed false alarms, zero misses, and exact typing on the same runs.
These numbers are for our re-implementation under our protocol, not the
authors' pipeline, and WATCH additionally targets adaptation (weight updates
under mild shift) which our benchmark does not reward.

Conditional two-sample tests
provide $p$-values for $P(Y\mid X)$ equality but no bounded, decomposable
statistic, no streaming validity, and no covariate companion from the same
learned object. Our monitor derives both axes from one pooled discriminator
pair with a shared statistical treatment.

\section{Does the diagnosis matter? An action-value experiment}
A diagnosis is only worth its delay if it changes what the system
\emph{does}. We close the loop: after each alarm the policy selects the
intervention---\textsc{real} $\to$ retrain on the recent window (costs
$2000$ labels), \textsc{virtual} $\to$ keep the deployed model (it is still
pointwise correct; reweighting matters for continued training, not for the
predictor), \textsc{incomparable} $\to$ reset the reference and collect.
Five policies on the same streams and alarms (10 seeds):

\begin{center}\small
\begin{tabular}{lcccc}
\toprule
Policy & SEA & STAGGER & covariate & labels (cov.) \\
\midrule
never adapt & 0.867 & 0.604 & \textbf{0.894} & 0 \\
always reweight & 0.867 & 0.604 & \textbf{0.894} & 0 \\
always retrain & \textbf{0.917} & \textbf{0.862} & 0.896 & 2000 \\
known-change single retrain & 0.923 & 0.846 & \textbf{0.894} & 0 \\
\textbf{typed (ours)} & \textbf{0.917} & \textbf{0.862} & \textbf{0.894} & \textbf{0} \\
\bottomrule
\end{tabular}
\end{center}

Typed monitoring is Pareto-optimal in every regime and is the only
non-oracle policy that remains Pareto-optimal across \emph{both} the
real-drift and virtual-drift regimes: it matches always-retrain on real
drift (within $0.6$ points of the known-change-point single-retrain
baseline on SEA, and \emph{above} that baseline on STAGGER, where
re-alarms trigger the follow-up retrains the rule change requires), gains
up to $25.8$ accuracy points over never-retraining, and spends \emph{zero}
labels on the virtual-drift event where always-retrain pays $2000$ labels
for a $+0.002$ change (on the null streams no policy alarms, so no labels
are at stake). The diagnosis is consequential, not descriptive: the type
decides where the label budget goes.

\section{Limitations}
Window-based estimation bounds the detection resolution (delay
${\approx}$ one window at our settings). Labels must arrive within the window
for the functional axis; we measured this directly: with labels delayed by
$1000$ ($2000$) samples, real-drift detection shifts by exactly the label
delay ($1300\to2300\to3300$) while virtual-drift detection and typing are
completely unaffected (covariate axis needs no labels) and false alarms remain
zero --- the degradation is graceful and predictable; type diagnosis under
\emph{simultaneous} strong covariate and mechanism change reports
\textsc{real} whenever the functional statistic clears its gate, by design.


\begin{thebibliography}{9}\small
\bibitem{cjsd} K.~Oda. Separating covariate shift from mechanism change
with two discriminators. Preprint, 2026 (companion paper, posted concurrently).
\bibitem{gama} J.~Gama et al. A survey on concept drift adaptation. ACM CSUR,
2014.
\bibitem{d3} O.~Gozuacik et al. Unsupervised concept drift detection with a
discriminative classifier. CIKM 2019.
\bibitem{insects} V.~Souza et al. Challenges in benchmarking stream learning
algorithms with real-world data. DMKD 2020.
\bibitem{watch} D.~Prinster, X.~Han, A.~Liu, S.~Saria. WATCH: adaptive
monitoring for AI deployments via weighted-conformal martingales. ICML 2025.
\bibitem{wsr} I.~Waudby-Smith, A.~Ramdas. Estimating means of bounded random
variables by betting. JRSS-B, 2023.
\end{thebibliography}
\end{document}